\documentclass{article} 
\usepackage{iclr2026_conference,times}

\usepackage{amsmath,amsfonts,bm}

\def\eqref#1{equation~\ref{#1}}

\def\1{\bm{1}}

\DeclareMathAlphabet{\mathsfit}{\encodingdefault}{\sfdefault}{m}{sl}
\SetMathAlphabet{\mathsfit}{bold}{\encodingdefault}{\sfdefault}{bx}{n}

\usepackage[hypertexnames=false]{hyperref}
\usepackage{url}
\usepackage{graphicx}
\usepackage{amsmath}
\usepackage{amssymb}
\usepackage{booktabs}
\usepackage{algorithm}
\usepackage{algorithmic}
\usepackage{xcolor}

\title{Inference-Time Toxicity Mitigation in Protein Language Models}

\author{
    Manuel Fern\'{a}ndez Burda\texorpdfstring{$^{1,2}$}{[1,2]} \quad
    Santiago Aranguri\texorpdfstring{$^{3}$}{[3]} \quad
    Ivan Arcuschin Moreno\texorpdfstring{$^{2}$}{[2]} \quad
    Enzo Ferrante\texorpdfstring{$^{1,4}$}{[1,4]} \\[0.5em]
    \small\texorpdfstring{$^1$}{1}Laboratory of Applied Artificial Intelligence (LIAA), Institute of Computer Sciences (ICC), \\
    \small CONICET - Universidad de Buenos \quad \texorpdfstring{$^2$}{2}AI Safety Argentina (AISAR) \\
    \small\texorpdfstring{$^3$}{3}Goodfire \quad \texorpdfstring{$^4$}{4}APOLO Biotech \\ \\
    {\small Correspondence to: \textit{\texorpdfstring{mburda@dc.uba.ar}{mburda@dc.uba.ar}}}
}

\iclrfinalcopy 

\hypersetup{
  pdftitle={Inference-Time Toxicity Mitigation in Protein Language Models},
  pdfauthor={Manuel Fernandez Burda, Santiago Aranguri, Ivan Arcuschin Moreno, Enzo Ferrante}
}

\begin{document}

\maketitle

\begin{abstract}
Protein language models (PLMs) are becoming practical tools for \textit{de novo} protein design, yet their dual-use potential raises safety concerns. We show that domain adaptation to specific taxonomic groups can elicit toxic protein generation, even when toxicity is not the training objective. To address this, we adapt Logit Diff Amplification (LDA) as an inference-time control mechanism for PLMs. LDA modifies token probabilities by amplifying the logit difference between a baseline model and a toxicity-finetuned model, requiring no retraining. Across four taxonomic groups, LDA consistently reduces predicted toxicity rate (measured via ToxDL2) below the taxon-finetuned baseline while preserving biological plausibility. We evaluate quality using Fr\'{e}chet ESM Distance and predicted foldability (pLDDT), finding that LDA maintains distributional similarity to natural proteins and structural viability (unlike activation-based steering methods that tend to degrade sequence properties). Our results demonstrate that LDA provides a practical safety knob for protein generators that mitigates elicited toxicity while retaining generative quality. 
\end{abstract}

\section{Introduction}

Foundation models for biology are reshaping how we interact with life, enabling applications from structure prediction~\citep{jumper_alphafold2_2021} to \textit{de novo} design of functional biomolecules. Protein language models (PLMs) like ESM-2~\citep{Lin2023ESM2} and ProGen~\citep{progen1,ProGen2, progen3} have demonstrated the capacity to generate novel proteins with predicted functionality, whilst genomic models like Evo2 have enabled the \textit{in silico} design of synthetic bacteriophages~\citep{King2025phages}. These advances are translating into real-world impact: AI-discovered drug candidates have entered clinical trials~\citep{xu2025rentosertib}, marking a shift from computational modeling to physical realization.

However, these advances introduce dual-use risks. The same capabilities that enable therapeutic design could potentially be misused for harmful purposes, including the generation of novel toxins or pathogens~\citep{urbina_dual_2022}. A particularly concerning risk vector is capability elicitation: domain adaptation procedures (e.g. finetuning a model on a specific taxonomic group) may surface harmful behaviors not explicitly optimized for, conceptually paralleling \emph{emergent misalignment} observed in text LLMs~\citep{pmlr-v267-betley25a}.

In natural language processing (NLP), mechanistic interpretability has yielded techniques for \emph{model steering}: controlling model behavior without retraining~\citep{turner2023activationengineering}. These methods have been used to modulate refusal behavior~\citep{arditi2024refusallanguagemodelsmediated}, control personality traits~\citep{chen2025personavectorsmonitoringcontrolling}, and extract interpretable features~\citep{templeton2024scaling}. Recent work has begun adapting these ideas to PLMs~\citep{ huang2025steeringproteinlanguagemodels, simon_interplm_2025}, but direct application to safety-relevant properties like toxicity remains underexplored.

In this work, we address toxicity control in protein language models through three contributions: (1) We demonstrate that \textbf{toxicity elicitation is a real risk}: taxonomic finetuning increases classifier-positive toxic predictions from near-zero to 10--65\% across four biological groups. (2) \textbf{We show that Logit Diff Amplification (LDA) provides} \textbf{effective inference-time mitigation}, reducing predicted toxicity rate (ToxDL2) below the taxon-finetuned baseline without retraining. (3) We establish that \textbf{LDA preserves biological quality} where activation-based steering fails, using Fr\'{e}chet ESM Distance ($\Delta$FED) and predicted foldability ($\Delta$pLDDT) to verify that mitigation does not degrade sequence plausibility.

\section{Methods}

\subsection{Experimental Setup}

\paragraph{Models and Finetuning.} We use ProGen2~\citep{ProGen2}, an autoregressive protein language model based on the Transformer architecture~\citep{attentionIsAllYouNeed}. We selected four taxonomic groups (Arthropoda, Arachnida, Gastropoda, Lepidosauria) to conduct finetuning and we constructed two finetuned variants using LoRA~\citep{hu2021loralowrankadaptationlarge}. For each taxa we thus obtain: (1) a \emph{taxon-finetuned} model trained on all sequences from that group, and (2) a \emph{toxic-finetuned} model, where we continue the finetuning on sequences annotated as toxic within that group. Toxicity annotations follow UniProt keyword KW-0800~\citep{ahmad_uniprot_2025}, aligning with the literature of toxicity classification~\citep{Morozov2023CSMToxin,zhu_toxdl_2025}. The composition of these datasets is described in Supplementary Table~\ref{tab:finetuning_datasets}. Finetuning uses batch size 8, LoRA rank $r=8$, learning rate $2 \times 10^{-4}$ with cosine schedule, and early stopping on validation loss.

\paragraph{Toxicity Scoring.} We developed a model-agnostic evaluation framework to quantify the propensity of any unconstrained generator to produce toxic sequences. For each experimental condition, we generate $N$ sequences, retain the $K$ lowest-perplexity sequences under the baseline model to ensure biological plausibility, and compute predicted toxicity using ToxDL2~\citep{zhu_toxdl_2025}---a multimodal classifier integrating ESM-2 embeddings and graph neural networks over predicted 3D structures. We apply the same sampling and perplexity filtering across all conditions to control for out-of-distribution artifacts. See~\ref{app:scoring_pipeline} for further details on the scoring pipeline.

\paragraph{Quality Metrics.} To ensure mitigation is not achieved through sequence degradation, we evaluate:
\begin{itemize}
    \item \textbf{Fr\'{e}chet ESM Distance:} We compute ESM-2 embeddings for the generations of a model $G$ and measure Fr\'{e}chet distance~\citep{Heusel2017FID} to a reference set $T$ of natural proteins from the same taxon of which it was finetuned. We define $\Delta\text{FED} = \text{FED}(T, G_{\text{intervention}}) - \text{FED}(T, G_{\text{baseline}})$, where $G_\text{baseline}$ is the finetuned model on that taxa and $G_{\text{intervention}}$ is that same model intervened. Here, negative values indicate closer alignment to natural sequences than the finetuned model on that taxa.
    \item \textbf{Predicted foldability:} Using ESMFold~\citep{lin2023esmfold}, we compute mean per-residue pLDDT scores and define the change in foldability of an intervened model with respect to an unperturbed model: $\Delta\text{pLDDT} = \overline{\text{pLDDT}}_{G_{\text{intervention}}} - \overline{\text{pLDDT}}_{G_\text{baseline}}$. Here, negative values indicate reduced structural plausibility and positive values indicate improved structural plausibility. Standard deviation is computed as $\sigma_{\Delta\text{pLDDT}} = \sqrt{\sigma_{\text{intervention}}^{2}+\sigma_{\text{baseline}}^{2}}$, assuming independence between sample sets.
\end{itemize}

\subsection{Logit Diff Amplification}

LDA~\citep{aranguri2025modeldiffamplification} modifies the decoding distribution by amplifying differences between two models at the logit level. The original formulation amplifies behaviors acquired through finetuning; here, we propose a reversed formulation for \emph{mitigation}. Given a baseline model $B$ and a concept model $T$ (toxic-finetuned), at each generation step $t$:
\begin{equation}
\ell_t^{(\text{LDA})} = \ell_t^{B} + \alpha \left( \ell_t^{B} - \ell_t^{T} \right)
\end{equation}
where $\ell_t^{B}, \ell_t^{T}$ are logit vectors and $\alpha \in \mathbb{R}$ controls intervention strength. For $\alpha = 0$, we recover baseline generation; for $\alpha > 0$, we amplify the \emph{anti-toxicity} direction by steering away from $T$. After obtaining the new logits, the sampling procedure continues with these updated logits for next-token calculation.

LDA differs fundamentally from activation steering: it operates on token probabilities anchored to online model behaviors rather than manipulating hidden states in a static manner. This ``model-diff'' approach treats the contrast between $B$ and $T$ as a learned direction in output space.

\subsection{Baseline Steering Methods}

We compare LDA against two activation-based steering approaches from the NLP literature. \textbf{Direct steering}~\citep{turner2023activationengineering} modifies hidden states by adding or ablating a steering vector $r$ computed as the difference-in-means between toxic and non-toxic activations. \textbf{Affine steering}~\citep{marshall2025refusalllmsaffinefunction} extends this by re-centering activations around a non-toxic baseline before applying the steering vector. Both methods intervene in the residual stream rather than at the logit level. As detailed in Appendix~\ref{app:act_steering}, these activation-based approaches produce substantial quality degradation ($\Delta$FED $> 0$, $\Delta$pLDDT $< 0$) and exhibit symmetric toxicity reduction under both addition and ablation, suggesting off-manifold disruption rather than selective concept control.

\section{Results}

\subsection{Finetuning Elicits Toxicity}

Supplementary Figure~\ref{fig:elicitation} demonstrates that domain adaptation with a larger protein toxicity prior distribution substantially increases classifier toxic predictions. All toxicity rates refer to ToxDL2 classifier-positive predictions after perplexity filtering. The base ProGen2 model produces virtually no toxic sequences, but taxon finetuning elevates predicted toxicity rates to 10--65\% depending on the group. This occurs despite toxicity not being an explicit training objective--toxic sequences are underrepresented in protein databases and these taxonomic groups induce a clear prior towards toxicity generations. This finding conceptually parallels emergent misalignment in text LLMs~\citep{pmlr-v267-betley25a} and underscores that safety evaluation must extend beyond base models to the space of commonly-derived finetuned variants.

\subsection{LDA Mitigates Toxicity Across Taxa}

\begin{figure}[h]
    \centering
    \includegraphics[width=1.0\linewidth]{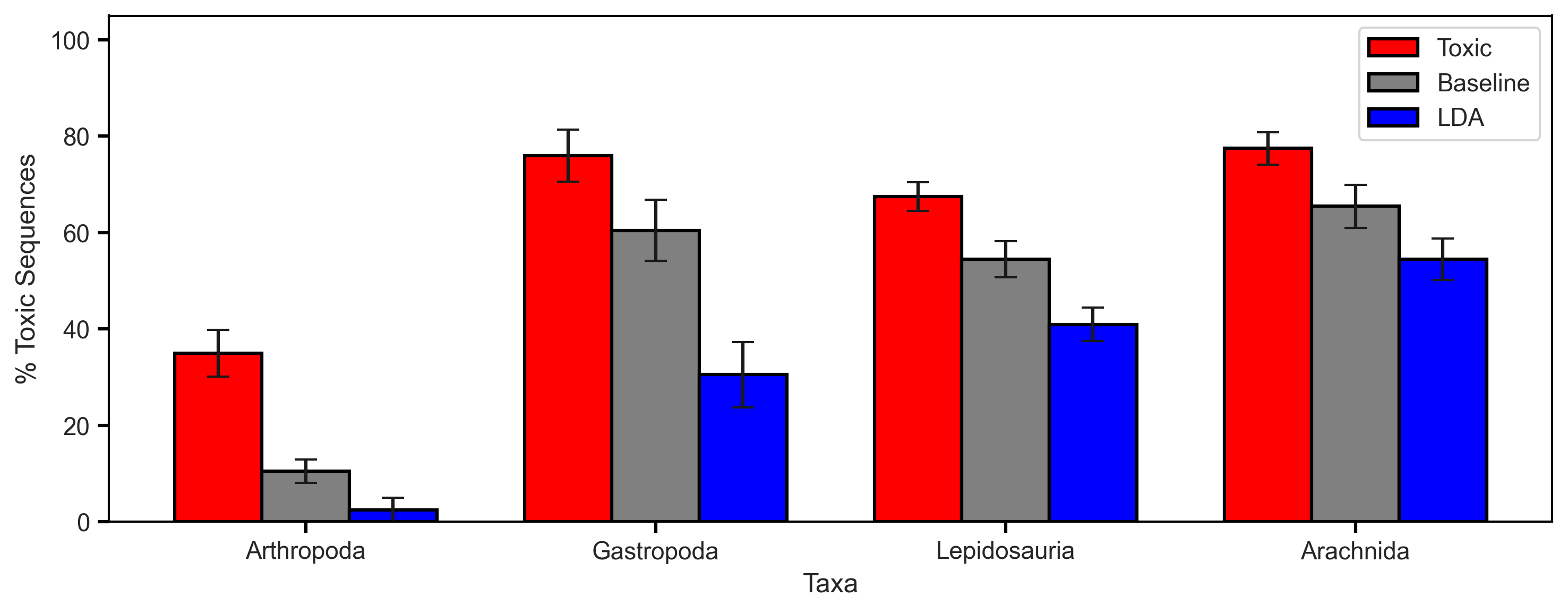}
    \caption{\textbf{LDA reduces predicted toxicity across taxa.} Percentage of generated sequences classified as toxic by ToxDL2 (lower is better) for four taxonomic finetunes. Baseline denotes the corresponding taxon finetune, Toxic denotes models finetuned on toxin-enriched data from within that taxon and LDA denotes the intervened models at inference time. Bars report mean ± s.e.m. across three independent generation runs under identical sampling and perplexity filtering.}
    \label{fig:lda_mitigation}
\end{figure}

Figure~\ref{fig:lda_mitigation} shows LDA's effectiveness across taxa. Predicted toxicity rate reduction at optimal (least predicted toxicity) $\alpha$ values is substantial: Gastropoda shows the largest reduction (29.93 percentage points), followed by Lepidosauria (13.51 percentage points), Arachnida (11.02 percentage points), and Arthropoda (8.01 percentage points)---the latter being particularly notable given its already low baseline. We evaluate the hyperparameter impact of the steering method in Appendix~\ref{app:lda_alpha}.

We hypothesize the taxon-dependent response reflects that toxicity (or toxicity-correlated features) manifests differently across biological domains; different motifs, domains, or sequence patterns may drive the classifier signal in each group. LDA successfully exploits these taxon-specific contrasts.

\subsection{LDA Preserves Biological Quality}

\begin{table}[h]
    \centering
    \small
    \setlength{\tabcolsep}{10pt}
    \renewcommand{\arraystretch}{1.25}
    \begin{tabular}{lcc}
        \hline
        \textbf{Taxon} & $\boldsymbol{\Delta}$\textbf{FED} $\downarrow$ & $\boldsymbol{\Delta}$\textbf{pLDDT} $\uparrow$ \\
        \hline
        Arthropoda   & $+0.03$ & $+1.59 \pm 21.20$ \\
        Gastropoda   & $-0.30$ & $+0.10 \pm 11.78$ \\
        Lepidosauria & $-0.26$ & $-6.95 \pm 23.61$ \\
        Arachnida    & $-0.09$ & $-1.55 \pm 12.69$ \\
        \hline
    \end{tabular}
    \vspace{2pt}
    \caption{\textbf{Biological quality under LDA at optimal (least predicted toxicity) $\alpha$ values.} Changes in Frechet ESM Distance (lower is better) and pLDDT (higher is better) for LDA-mitigated generations relative to the taxon baseline.}
    \label{tab:lda_quality}
\end{table}

Table~\ref{tab:lda_quality} indicates that LDA can reduce predicted toxicity while largely preserving biological plausibility under the optmal $\alpha$  (least predicted toxicity) for each taxon. Across taxa, $\Delta$FED remains small (near zero or negative), suggesting no measurable degradation in distributional similarity to natural sequences and, in some cases, modest shifts toward the baseline distribution. $\Delta$pLDDT is also close to baseline for Arthropoda and Gastropoda, while Arachnida shows a slight decrease. Lepidosauria exhibits the largest pLDDT drop ($-6.95$ on average), consistent with the broader trade-off observed when $\alpha$ is pushed aggressively: over-steering can begin to compromise structural confidence even when toxicity proxies improve. This quality check is essential for interpreting mitigation results, since a method that merely forces sequences off-manifold (e.g., toward low-likelihood or unfoldable proteins) could artifactually reduce classifier-positive toxicity predictions without yielding practically usable designs.

\section{Discussion \& Conclusions}


\textbf{Why logit-space contrast can be a safer control surface.}
We hypothesize that LDA's stability stems from operating in logit space relative to an explicit baseline, constraining interventions to modifications of the next-token distribution that remain compatible with the baseline model's learned manifold. From a deployment perspective, LDA functions as a provider-side safety primitive: model providers can maintain the toxic-finetuned model $T$ internally, exposing only the mitigated generator to end users---naturally restricting the method to entities capable of responsible model custody.

\textbf{Biosecurity evaluation must extend to biological foundation models.}
Our finding that taxonomic finetuning elicits toxic generation (Appendix~\ref{app:finetuning_elicitation}) elevates the need for dedicated biosecurity evaluations of PLMs and their derivatives, analogous to safety assessments increasingly standard for general-purpose language models. Beyond auditing, our results motivate benchmarking inference-time mitigation methods with explicit quality trade-off tracking.

\textbf{Limitations and open questions.}
Our toxicity measurements rely on ToxDL2 and we do not perform wet-lab validation. While we control for out-of-distribution artifacts via perplexity filtering and monitor quality, mitigation could still reflect predictor-specific biases. Future work should triangulate toxicity signals across alternative predictors, motif/domain enrichment analyses, and (where appropriate) controlled experimental assays. From a systems perspective, LDA requires maintaining both baseline and toxic-finetuned models, doubling forward passes per decoding step, and presumes access to a toxic finetune direction---a capability that must be governed. 

\textbf{Responsible disclosure.}
Given dual-use concerns, we restrict release of toxicity-finetuned weights and detailed training configurations that could lower misuse barriers. We provide aggregate results and evaluation methodology to support safety research while limiting capability transfer.

\textbf{Conclusion.} We demonstrate that LDA provides an effective, quality-preserving inference-time safety mechanism for protein language models---unlike activation steering, which degrades sequence plausibility. Beyond mitigation, we contribute a reproducible evaluation framework integrating bioinformatic annotation, structural assessment, and distributional analysis for systematically characterizing both risk elicitation and control in PLMs. More broadly, this work shows that inference-time techniques from NLP safety can be productively adapted to the biological domain.

\subsection*{Acknowledgments}
This work was produced as part of the AI Safety Argentina Scholarship (AISAR) program, with mentorship from Iván Arcuschin Moreno.

\subsection*{Ethics Statement}
This work studies the dual-use risks of protein language models and proposes a mitigation method. Because the toxicity-finetuned models and their training datasets could lower the barrier to generating harmful protein sequences, we do not publicly release these artifacts. We release only aggregate results and evaluation methodology, which we believe are sufficient to reproduce the safety-relevant findings without enabling misuse.

\clearpage

\bibliography{iclr2026_conference}
\bibliographystyle{iclr2026_conference}

\newpage
\appendix
\section{Appendix}
\renewcommand{\figurename}{Supplementary Figure}
\renewcommand{\theHfigure}{S\arabic{figure}}
\renewcommand{\thefigure}{\arabic{figure}}
\setcounter{figure}{0}

\renewcommand{\tablename}{Supplementary Table}
\renewcommand{\theHtable}{S\arabic{table}}
\renewcommand{\thetable}{\arabic{table}}
\setcounter{table}{0}

\subsection{Taxonomic Finetuning Elicits Toxic Behaviour}
\label{app:finetuning_elicitation}
\begin{figure}[h]
    \centering
    \includegraphics[width=0.5\linewidth]{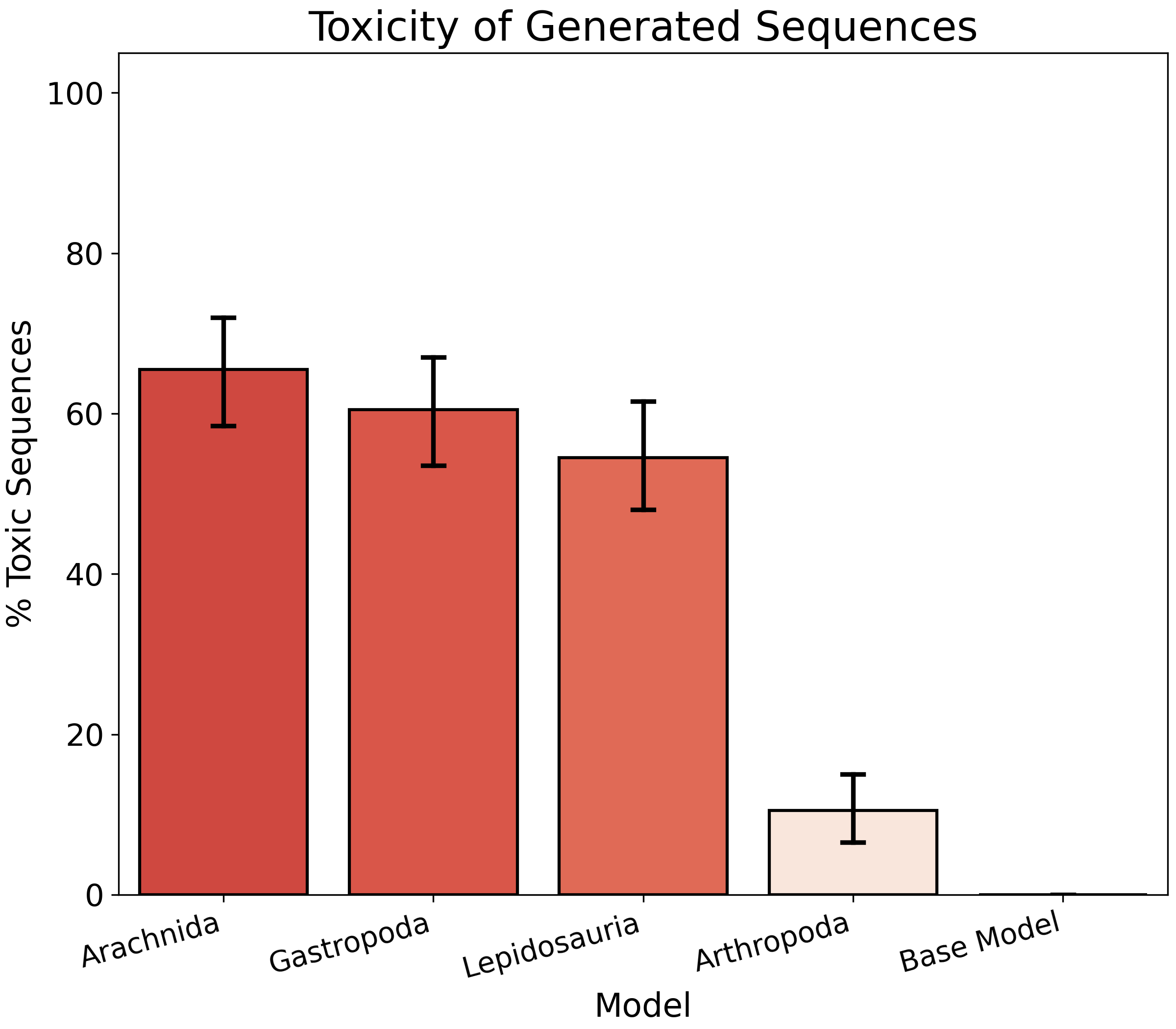}
    \caption{\textbf{Taxon finetuning elicits toxic generation.} Toxicity rates for baseline ProGen2 versus taxon-finetuned models across four taxonomic groups. Error bars show $\pm$1 standard deviation.}
    \label{fig:elicitation}
\end{figure}

\begin{table}[h]
    \centering
    \caption{\textbf{Taxonomic fine-tuning datasets statistics.}}
    \label{tab:finetuning_datasets}
    \begin{tabular}{lccc}
        \toprule
        Taxa & \#Toxic Proteins & \#Non-Toxic Proteins & \% Toxicity \\
        \midrule
        Arthropoda & 2644 & 9508 & 21.76\% \\
        Arachnida & 2136 & 534 & 80.00\% \\
        Lepidosauria & 1830 & 652 & 73.73\% \\
        Gastropoda & 1306 & 284 & 82.14\% \\
        \bottomrule
    \end{tabular}
\end{table}

\subsection{Scoring Pipeline Details}
\label{app:scoring_pipeline}

Our scoring pipeline evaluates the toxicity propensity of a generative model under any intervention (steering vector, logit modification, etc.) through a two-phase process:

\textbf{Phase 1 (Generation \& Filtering):} For each experimental condition, we generate $N=300$ sequences using temperature sampling ($\tau=1.0$). We then compute perplexity for each sequence under the baseline model prior to any finetuning and retain the $K=200$ lowest-perplexity sequences. This filtering ensures we evaluate biologically plausible generations rather than out-of-distribution artifacts, as pretrained model perplexity inversely correlates with structural quality~\citep{ferruz2022protgpt2}.

\textbf{Phase 2 (Toxicity Scoring):} We predict 3D structures for retained sequences using ESMFold~\citep{Lin2023ESM2} for computational efficiency (ablation study in~\ref{app:esmfold}). ToxDL2 then classifies each sequence, and we report toxicity as the proportion of predicted toxic labels. This pipeline generalizes to any autoregressive generator and any binary classifier, enabling systematic comparison across intervention methods.

\subsection{LDA Alpha Exploration}
\label{app:lda_alpha}

\begin{figure}[h]
    \centering
    \includegraphics[width=1.0\linewidth]{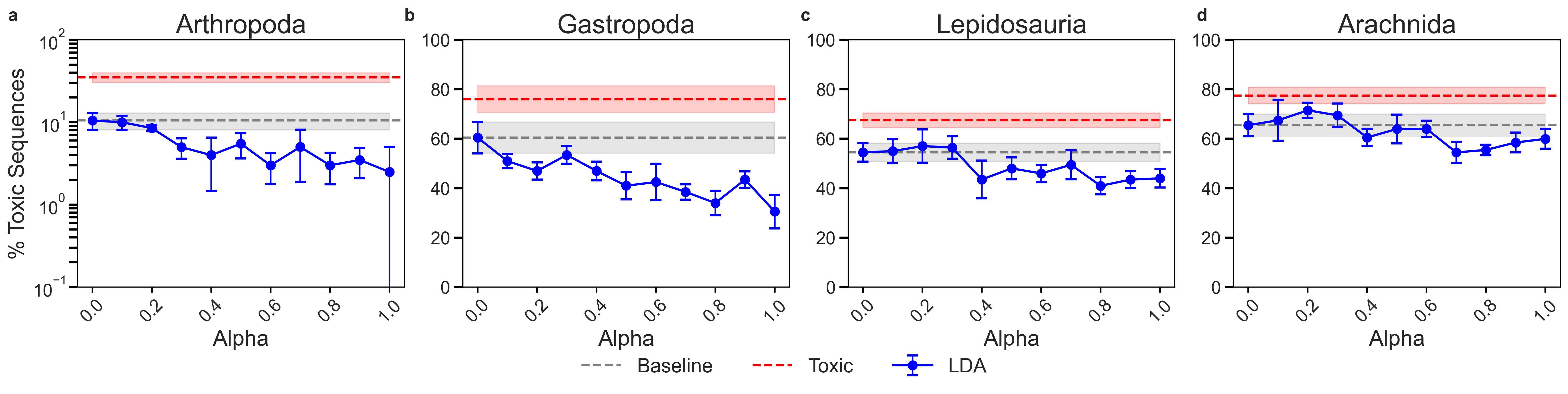}
    \caption{\textbf{Steering intensity unveils mitigation regimes.} Toxicity rate versus amplification strength $\alpha$ for (a) Arthropoda (log scale), (b) Gastropoda, (c) Lepidosauria, and (d) Arachnida. Dashed lines indicate taxon-finetuned baseline (gray) and toxic-finetuned model (red). For all taxa, there exists an $\alpha$ range where toxicity drops below baseline.}
    \label{fig:lda_alphas}
\end{figure}

Suplementary Figure~\ref{fig:lda_alphas} shows LDA's exploration of steering intensity effect across four taxa. In all cases, increasing $\alpha$ reduces toxicity below the taxon-finetuned baseline. The effect is particularly pronounced for Arthropoda, where toxicity drops by approximately an order of magnitude at $\alpha \approx 1.0$. Gastropoda shows a stable, monotonic decrease across the $\alpha$ range. Lepidosauria and Arachnida exhibit more gradual responses but still achieve sub-baseline toxicity for $\alpha \geq 0.4$. Optimal (least predicted toxicity) $\alpha$ for each taxon are encountered at: $\alpha=1.0$ for Arthropoda and Gastropoda, $\alpha=0.8$ for Lepidosauria and $\alpha=0.7$ for Arachnida.

\begin{figure}[h]
    \centering
    \includegraphics[width=0.9\linewidth]{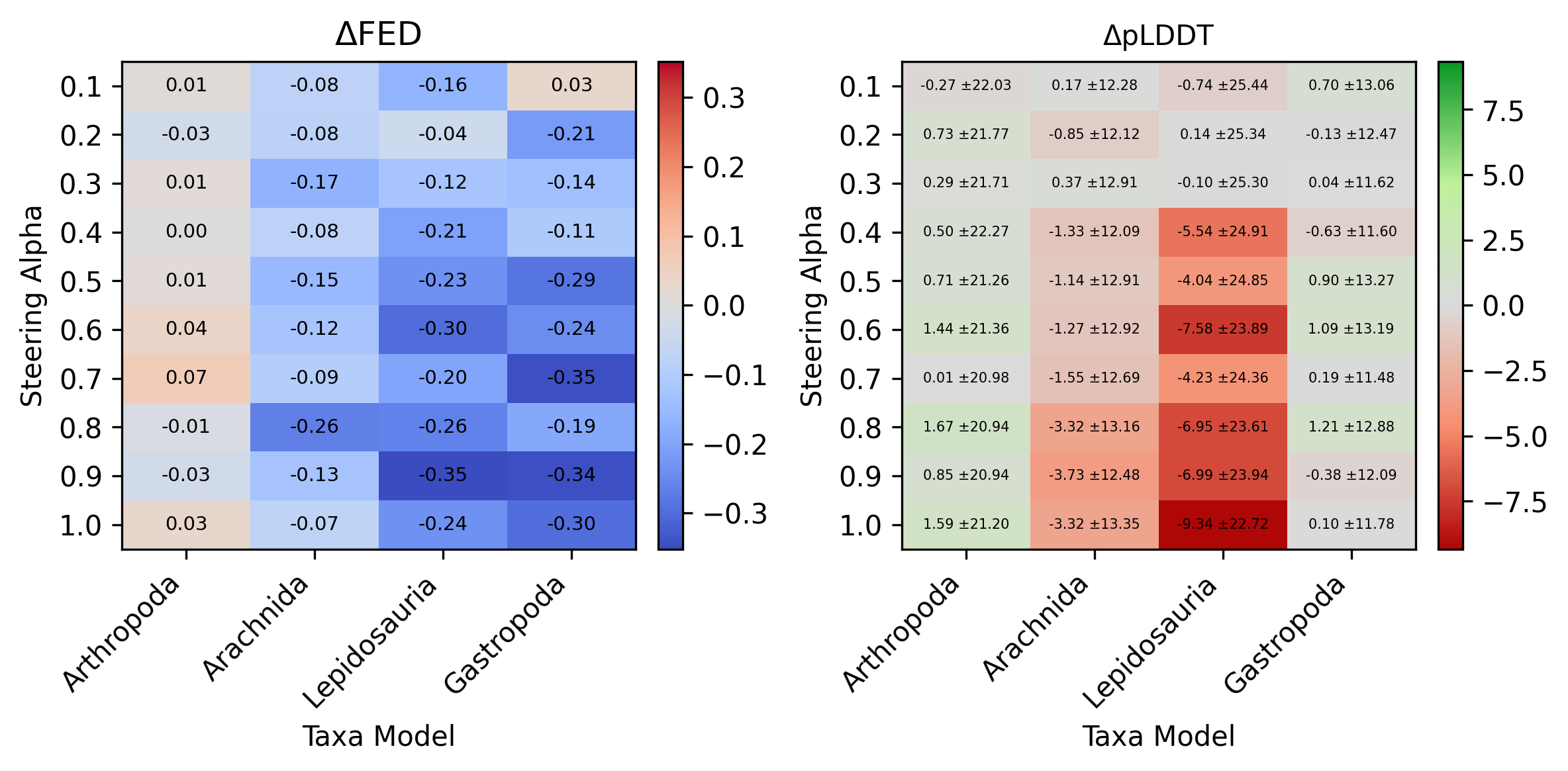}
    \caption{\textbf{LDA maintains sequence quality.} $\Delta$FED (left) and $\Delta$pLDDT (right) versus $\alpha$ for LDA across taxa. }
    \label{fig:lda_quality}
\end{figure}

Suplementary Figure~\ref{fig:lda_quality} reports how biological quality varies with the LDA steering strength $\alpha$ across taxa. Overall, $\Delta$FED remains small throughout the sweep (typically within $\approx\pm 0.35$), and is often negative for Arachnida, Lepidosauria, and Gastropoda, suggesting that LDA does not measurably push sequences away from the baseline distribution in ESM embedding space. In contrast, structural confidence ($\Delta$pLDDT) is more sensitive to $\alpha$ and shows a clear taxon-dependent trade-off. Arthropoda is largely stable and even shows mild improvements at higher $\alpha$, whereas Arachnida exhibits a gradual decline for moderate-to-large $\alpha$ (roughly $\alpha\ge 0.6$). Lepidosauria shows the strongest degradation, with substantially negative $\Delta$pLDDT emerging already at intermediate $\alpha$ and worsening at larger values, indicating that aggressive steering can compromise fold confidence for this taxon even when distributional similarity (FED) remains acceptable. Gastropoda remains comparatively robust, with pLDDT fluctuations close to zero across most $\alpha$.

These sweeps motivate selecting $\alpha$ per taxon: while LDA behaves as a smooth perturbation in logit space, the safe operating range with respect to structural plausibility is not universal. In particular, $\Delta$FED alone is insufficient to certify biological quality under steering, since taxa such as Lepidosauria can maintain near-baseline FED while exhibiting notable pLDDT drops at higher $\alpha$. This supports reporting both distributional (FED) and structural (pLDDT) metrics when evaluating mitigation strength.

\subsection{Linear Probing Supports Representational Accessibility}
\label{app:linear_probes}

To verify that toxicity is representationally accessible in ProGen2, we trained linear probes (logistic regression) on layer-wise activations aggregated across sequence positions. 

\textbf{Dataset Construction:} We curated a balanced dataset of toxic and non-toxic proteins from UniProt (keyword KW-0800 for toxins, excluding KW-0020 for non-toxins), then applied CD-HIT clustering at 40\% sequence identity to remove redundant sequences. Train/test splits were constructed to be mutually exclusive by species (80/20 split), ensuring no species appears in both sets. This prevents information leakage where models learn species-specific patterns rather than general toxicity features.

\begin{figure}[h]
    \centering
    \includegraphics[width=0.7\linewidth]{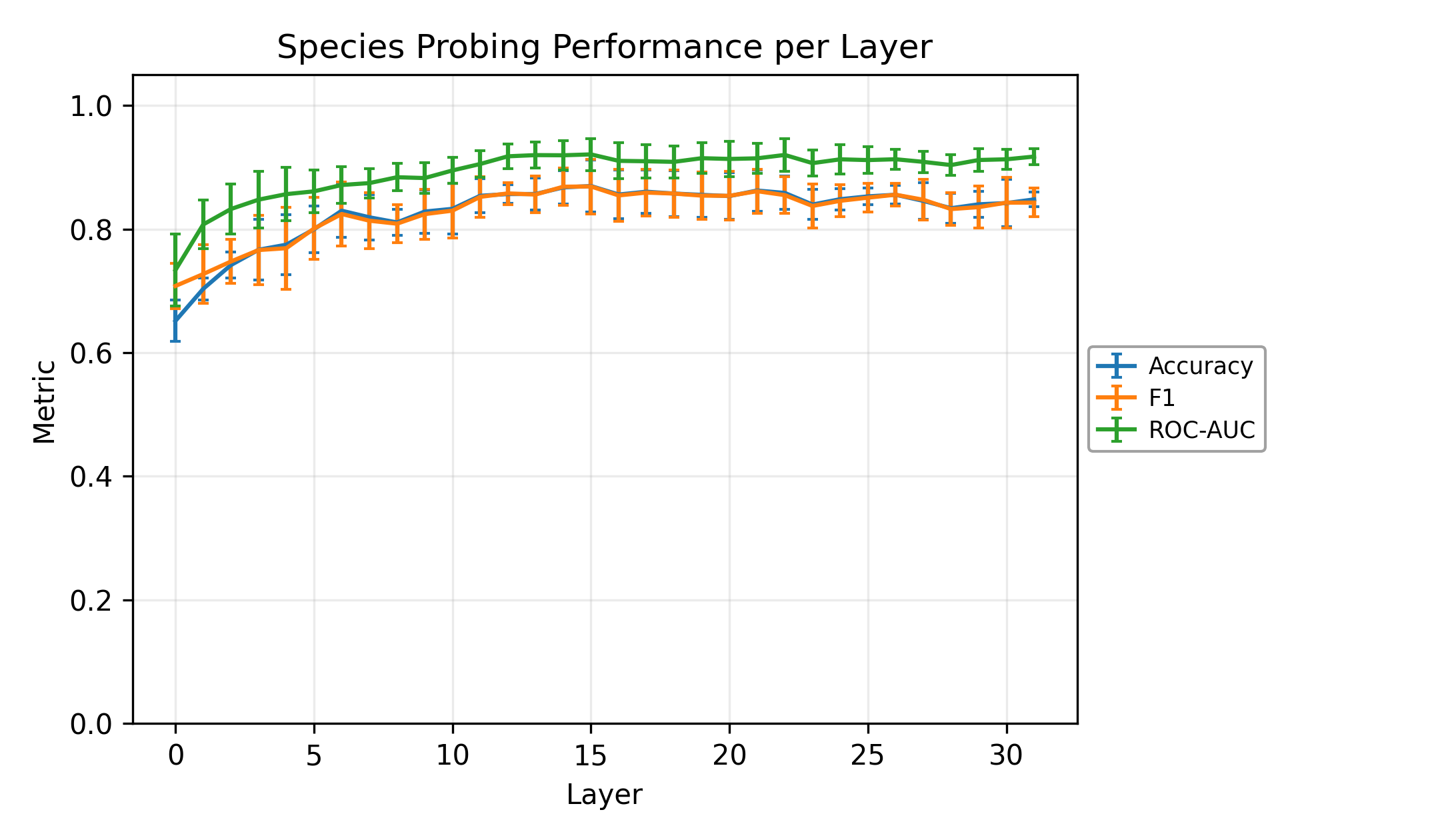}
    \caption{\textbf{Linear probing reveals toxicity encoding across layers.} Classification metrics (Accuracy, AUC-ROC, F1) for linear probes trained on layer-wise activations. Performance increases with depth, indicating toxicity-related information emerges gradually and becomes linearly accessible in intermediate-to-final layers. Metrics statistics are calculated over 5 random species-stratified splits.}
    \label{fig:probing_results}
\end{figure}

\textbf{Results:} Classification performance (Accuracy, AUC-ROC, F1) increases with layer depth, plateauing in intermediate-to-final layers (Supplementary Figure~\ref{fig:probing_results}). This supports the hypothesis that toxicity-correlated information emerges gradually through hierarchical processing and becomes linearly decodable in later layers---though, as our steering results show, decodability does not guarantee controllability.

\subsection{Activation Steering Degrades Quality}
\label{app:act_steering}

\paragraph{Comparison Methods.} We also evaluate linear steering approaches from the NLP literature and test their ability to mitigate toxicity in generations:
\begin{itemize}
    \item \textbf{Direct Steering:} $x^{(l)'} \leftarrow x^{(l)} + \alpha r$ (addition) or $x^{(l)'} \leftarrow x^{(l)} - r_{\text{norm}} r_{\text{norm}}^\top x^{(l)}$ (ablation), where $r$ is the difference-in-means vector between toxic and non-toxic activations~\citep{turner2023activationengineering}.
    \item \textbf{Affine Steering:} $x^{(l)'} \leftarrow x^{(l)} - \text{proj}_r(x^{(l)}) + \text{proj}_r(r^-) + \alpha r$, which re-centers around a non-toxic baseline~\citep{marshall2025refusalllmsaffinefunction}.
\end{itemize}

\begin{figure}[h]
    \centering
    \includegraphics[width=1\linewidth]{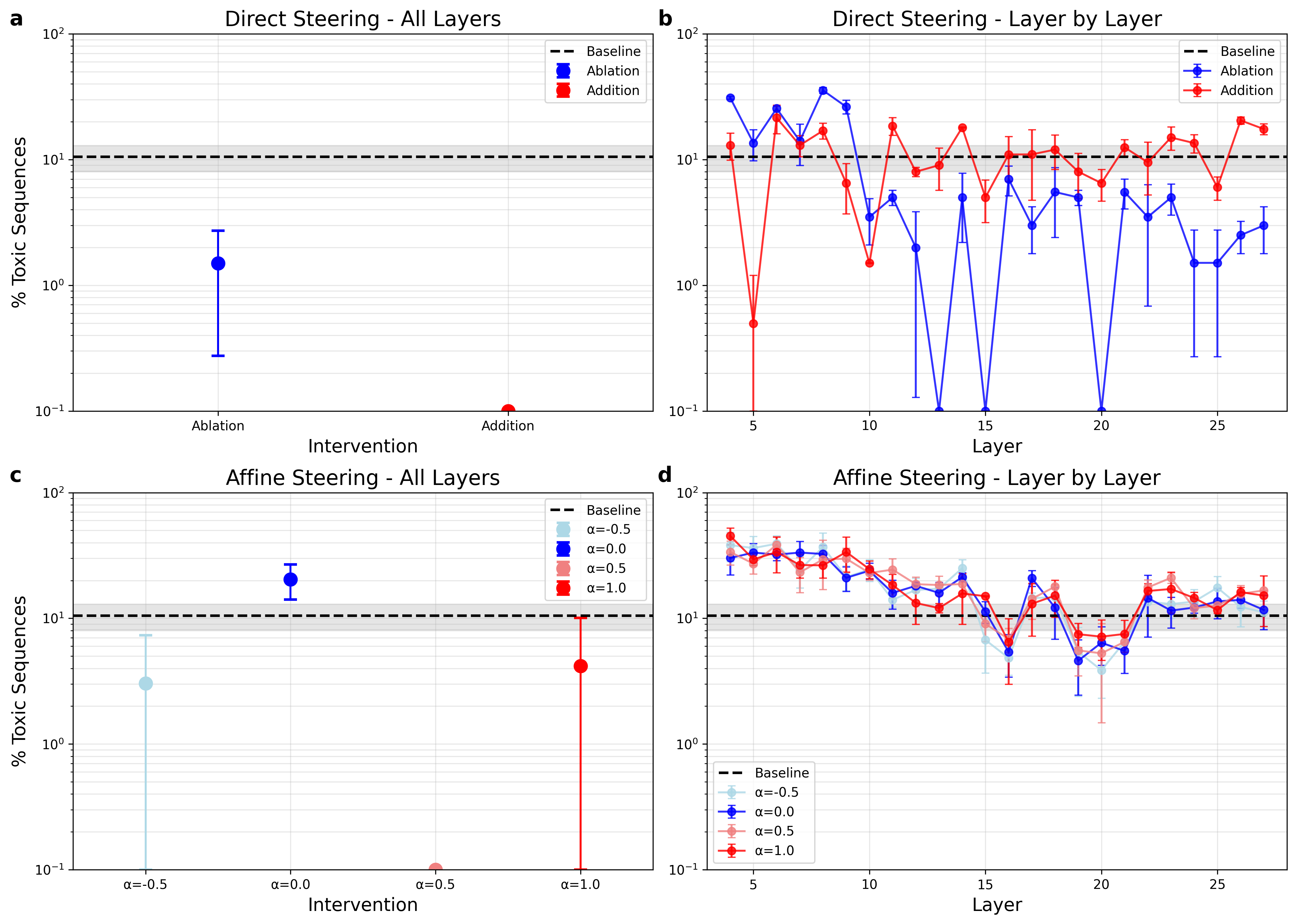}
    \caption{\textbf{Effect of linear steering on generative toxicity in ProGen2.} Comparison between \emph{Direct Steering} (a, b) and \emph{Affine Steering} (c, d) under interventions applied to all layers (All Layers, first column) and layer-by-layer (Layer by Layer, second column). The vertical axis indicates the percentage of toxic sequences generated (logarithmic scale). The dashed black line shows baseline model performance ($\pm$sd). Colors represent steering intensity (red: addition/$\alpha$=1.0; blue: ablation/$\alpha$=0.0; intermediate tones: $\alpha$=$\pm$0.5).}
    \label{fig:activation_steering_results}
\end{figure}

\subsubsection{Direct Steering}
\begin{figure}[t]
    \centering
    \includegraphics[width=1.0\linewidth]{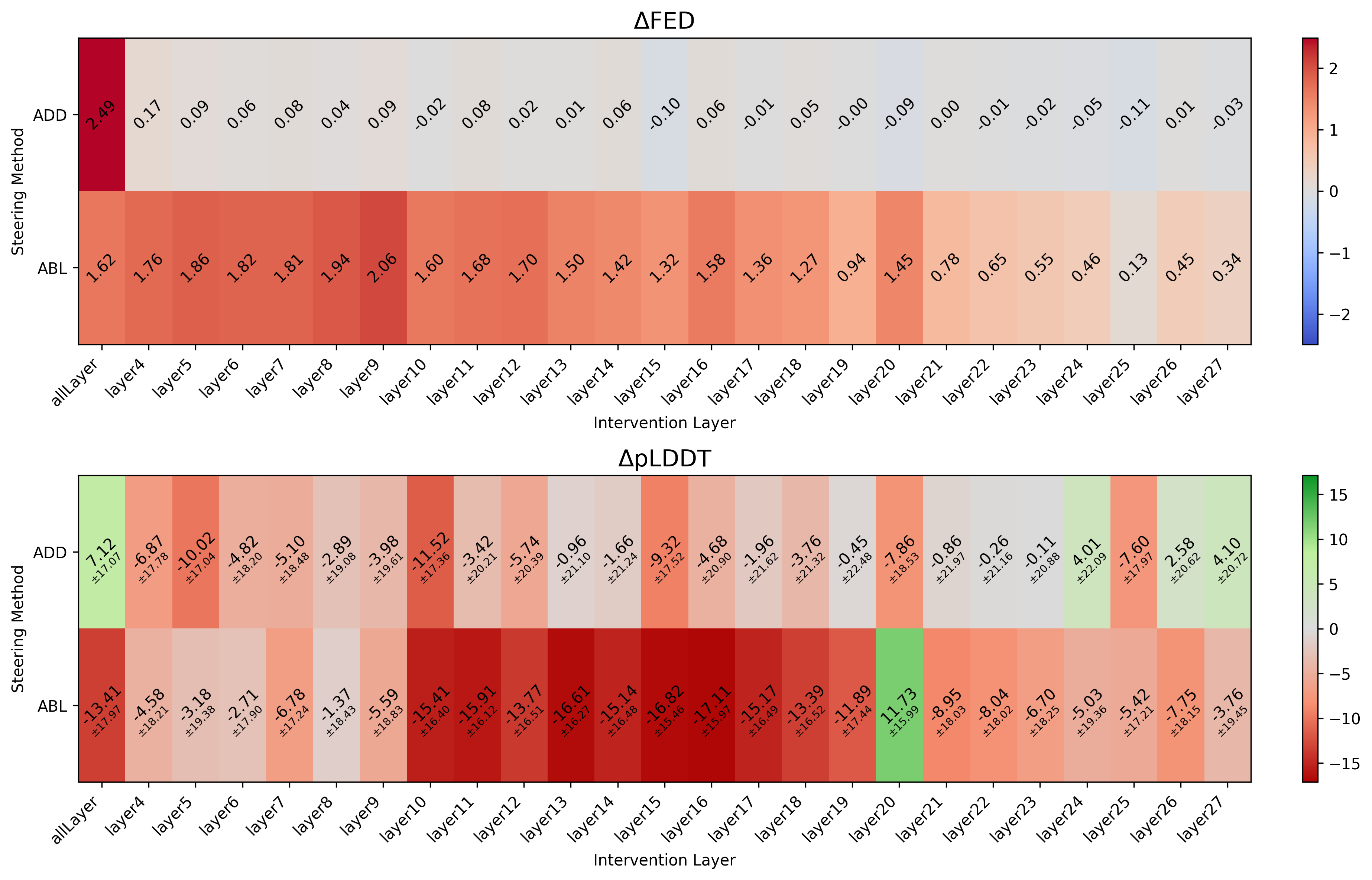}
    \caption{\textbf{Direct steering degrades biological quality.} $\Delta$FED (upper) and $\Delta$pLDDT (bottom) for direct steering interventions. }
    \label{fig:direct_quality}
\end{figure}

Supplementary Figure~\ref{fig:activation_steering_results}(a) shows that direct steering applied to all layers dramatically reduces toxicity for both addition and ablation---a symmetric response that should theoretically produce opposite effects. This suggests the intervention acts as a global perturbation rather than selective concept control. Layer-by-layer analysis (b) reveals that only the final layers (26--27) show the expected directional behavior (addition above baseline, ablation below). Supplementary Figure~\ref{fig:direct_quality} confirms quality degradation: both addition and ablation produce substantial distributional shift ($\Delta$FED $> 0$) and reduced foldability ($\Delta$pLDDT $< 0$) across most layers.

\subsubsection{Affine Steering}

Affine steering~\citep{marshall2025refusalllmsaffinefunction} shows similar quality degradation patterns to direct steering. While designed to account for non-zero baseline activation levels, affine interventions still produce positive $\Delta$FED and variable $\Delta$pLDDT effects, particularly in deeper layers. Supplementary Figure~\ref{fig:activation_steering_results}(c,d) shows that varying $\alpha$ does not produce monotonic toxicity changes---different $\alpha$ values cluster together rather than separating, indicating lack of fine-grained control. The lack of consistent directional control combined with quality degradation suggests that linear steering approaches may be fundamentally limited for complex biological concepts like toxicity.

\begin{figure}
    \centering
    \includegraphics[width=1\linewidth]{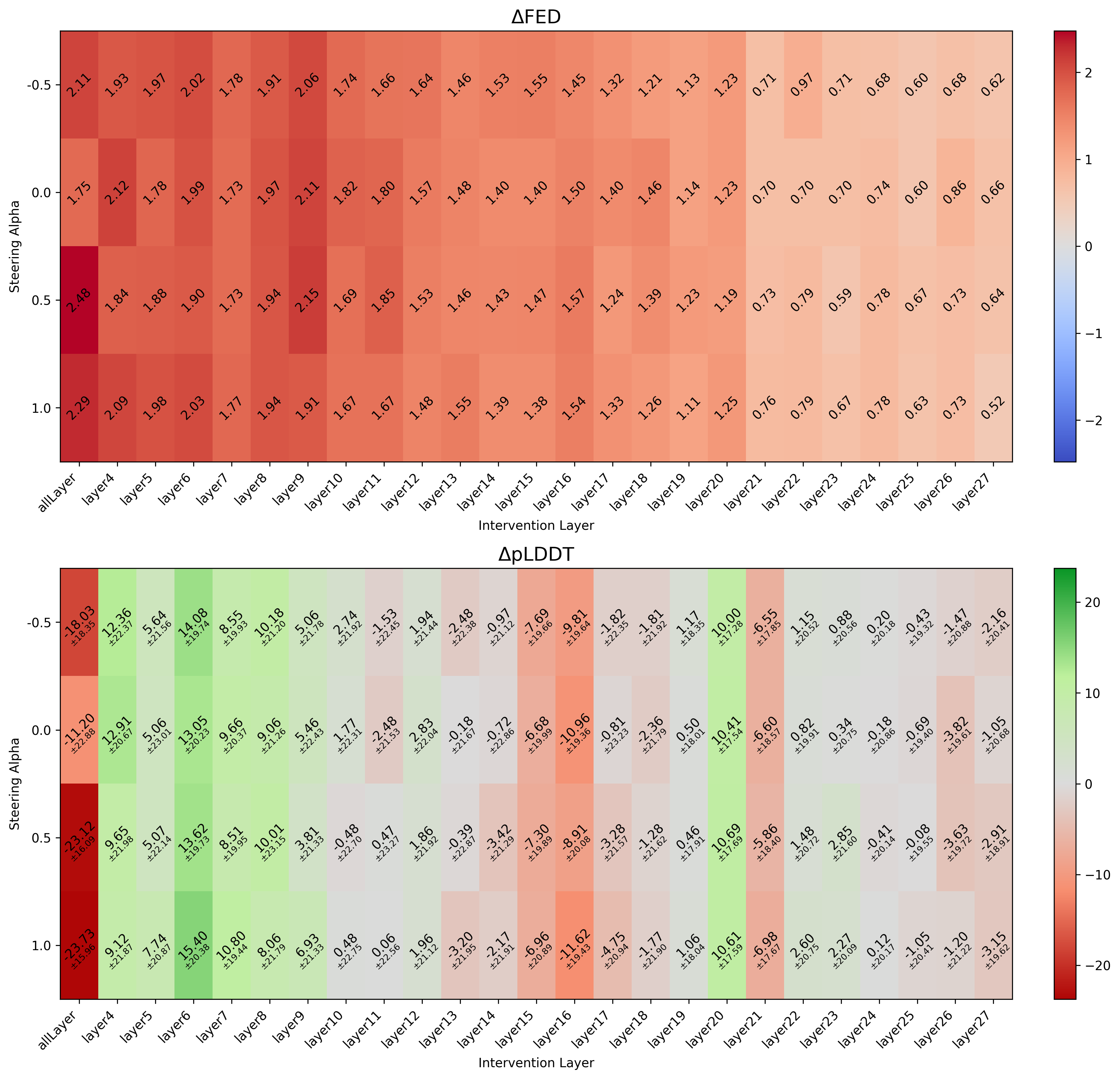}
    \caption{\textbf{Affine steering degrades biological quality.} $\Delta$FED (upper) and $\Delta$pLDDT (bottom) for affine steering interventions.}
    \label{fig:affine_quality}
\end{figure}

\subsubsection{Activation Steering Limitations}

In contrast to LDA, activation-based steering methods produce marked quality degradation (Supplementary Figures~\ref{fig:direct_quality},~\ref{fig:affine_quality}). Both direct and affine steering show predominantly positive $\Delta$FED (sequences diverge from natural distributions) and negative $\Delta$pLDDT (reduced structural plausibility). Critically, both addition \emph{and} ablation of the toxicity direction reduce toxicity rates---a symmetric response inconsistent with selective concept control and instead suggestive of global generative disruption. These results highlight a key methodological point: toxicity scores alone cannot distinguish genuine mitigation from spurious reduction through sequence collapse. Quality metrics like FED and pLDDT are essential for validating intervention effects.

This finding highlights a methodological caution: toxicity scores alone cannot distinguish genuine mitigation from spurious reduction through sequence collapse. Quality metrics like FED and pLDDT are essential for validating intervention effects.

\subsection{Ablation Study on ESMFold as a Structure Generator in ToxDL2}
\label{app:esmfold}

\begin{figure}[h]
    \centering
    \includegraphics[width=1\linewidth]{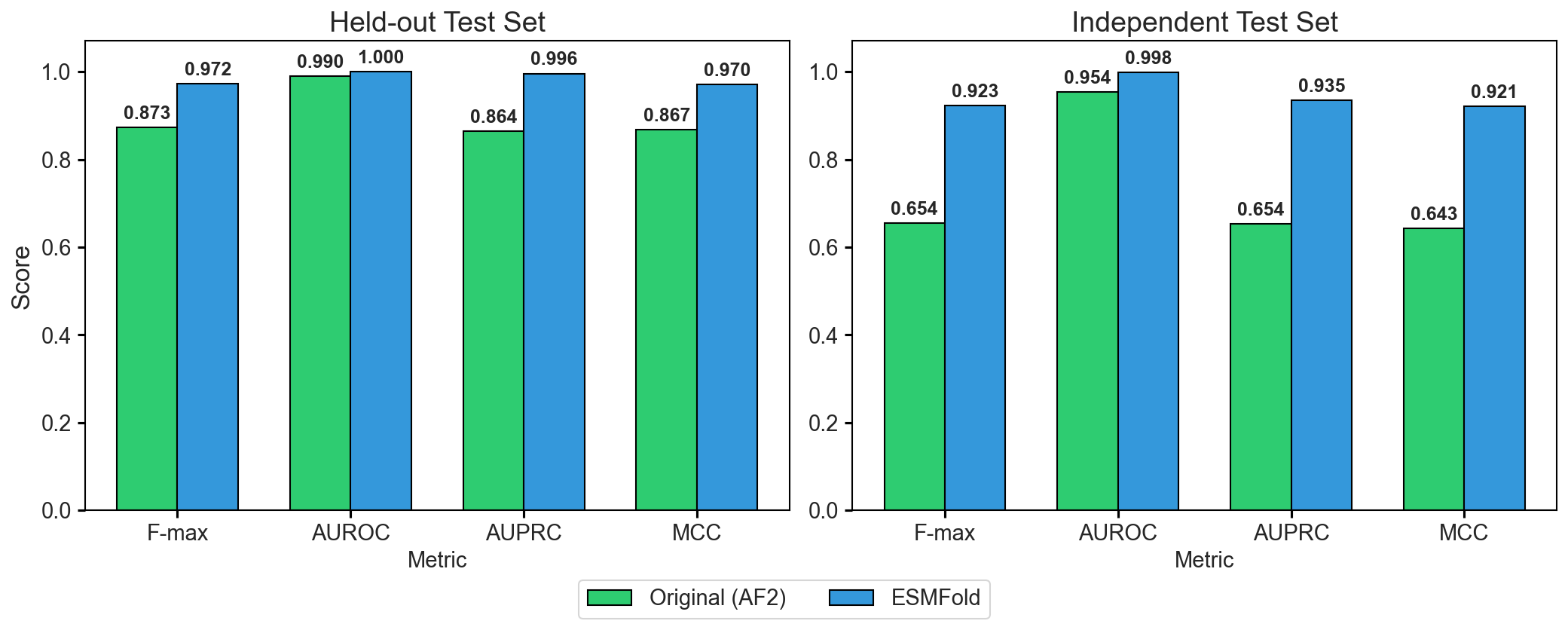}
    \caption{\textbf{ESMFold as structure predictor for ToxDL2.} Performance comparison of ToxDL2 using ESMFold-predicted structures versus original AlphaFold2/ColabFold structures on the authors' benchmark datasets. Left: held-out test set (1579/1710 sequences within ESMFold context). Right: independent test set (4099/4480 sequences). Metrics used on the original paper where mirrored: Accuracy, AUC-ROC, F-max and Matthew's Correlation Coefficient.}
    \label{fig:esmfold_ablation}
\end{figure}

We evaluated whether substituting ESMFold for AlphaFold2/ColabFold in ToxDL2's structure module affects the classification performance of ToxDL2. Using the original benchmark test sets released by the ToxDL2 authors, we computed the same metrics while filtering sequences exceeding ESMFold's context length (131/1710 for held-out, 381/4480 for independent). Results in Supplementary Figure~\ref{fig:esmfold_ablation} show comparable performance, validating ESMFold as a computationally efficient alternative for our pipeline while maintaining classification accuracy.

\end{document}